\def\year{2019}\relax
\documentclass[letterpaper]{article} 
\usepackage{aaai19}  
\usepackage{times}  
\usepackage{helvet}  
\usepackage{courier}  
\usepackage{url}  
\usepackage{graphicx}  
\usepackage{fancyvrb}
\usepackage{enumitem}
\usepackage{amssymb}
\usepackage{booktabs}
\usepackage[strings]{underscore}

\begin{document}
%
\title{Stepping Stones to Inductive Synthesis of Low-Level Looping Programs}
\author{Christopher D. Rosin \\
Parity Computing, Inc.\\
San Diego, California, USA \\
christopher.rosin@gmail.com}
\maketitle

\begin{abstract}
Inductive program synthesis, from input/output examples, can provide
an opportunity to automatically create programs from scratch without
presupposing the algorithmic form of the solution.  For induction of
general programs with loops (as opposed to loop-free programs, or
synthesis for domain-specific languages), the state of the art is at
the level of introductory programming assignments.  Most problems that
require algorithmic subtlety, such as fast sorting, have remained out
of reach without the benefit of significant problem-specific
background knowledge.  A key challenge is to identify cues that are
available to guide search towards correct looping programs.  We
present MAKESPEARE, a simple delayed-acceptance hillclimbing method
that synthesizes low-level looping programs from input/output
examples.  During search, delayed acceptance bypasses small gains to
identify significantly-improved stepping stone programs that tend to
generalize and enable further progress.  The method performs well on a
set of established benchmarks, and succeeds on the previously unsolved
``Collatz Numbers'' program synthesis problem.  Additional benchmarks
include the problem of rapidly sorting integer arrays, in which we
observe the emergence of comb sort (a Shell sort variant that is
empirically fast).  MAKESPEARE has also synthesized a record-setting
program on one of the puzzles from the TIS-100 assembly language
programming game.
\end{abstract}

\section{Introduction}

Automated synthesis of programs from user requirements has a long history as an AI research goal \cite{prowijcai69,gulwanibook}.
Recent interest in the problem has led to synthesis success for non-looping programs (e.g. clever bit-twiddling \cite{gulwaniloopfree}), 
partial program ``sketches'' with holes to be synthesized \cite{SKETCH,korea}, 
domain-specific languages (e.g. Flash Fill in Microsoft Excel \cite{flashfill}), and other areas.
But for synthesis of general looping programs from scratch, the state of the art is at the level
of introductory programming exercises \cite{spector,push2018,pldi2015}.
Most problems requiring algorithmic subtlety, such as fast sorting, have remained out of reach without using significant
problem-specific background knowledge \cite{dawnsong,evolvingefficientrecursivesortinglucas}.

We seek a simple method that can make progress and provide insight on synthesis problems requiring algorithmic subtlety.
We target low-level programming languages,
close to assembly language; this provides some grounding (avoiding
open-ended exploration of high-level language design), and has the potential to yield a large speed advantage.
We focus on inductive program synthesis\footnote{``Program synthesis'' here refers generally to the automated creation of programs that meet user-supplied specifications,
and {\em inductive} program synthesis indicates that the specification takes the form of input/output examples -- see \cite{difficulty} for similar usage.
Others have used the term program {\em induction} to indicate that the program is implicit, e.g. in a neural network \cite{robustfill}, instead of being explicitly generated. 
In this paper though, we seek to output an explicit program.}
 from input/output examples, which provides an opportunity 
to automatically create programs without presupposing the algorithmic form of the solution (other approaches such as
synthesis from natural language description \cite{cardgamebenchmark} or deduction from formal specifications \cite{deductive}
often constrain the form of the solution \cite{gulwanibook}).  Our goal is {\em pure inductive synthesis}:
without problem-specific primitives or background knowledge.

The search space of programs is vast, and a key challenge is to identify cues that are available to guide search towards correct programs.
We focus on stochastic search in the hopes of tackling programs that are prohibitively complex for exhaustive methods (we note though that work
on exhaustive methods is progressing \cite{deepcoder,korea}).  While problems requiring simple loops (e.g. same operation
on each array element) can provide a clear path for stochastic search to hillclimb,
problems requiring more complex loops could lack partially-correct {\em stepping stone} programs that
enable progress to complete solutions.

We find a simple {\em Delayed Acceptance} extension to hillclimbing
can identify stepping stones that lead to successful synthesis of looping programs with challenging requirements.
Our method, {\em MAKESPEARE}, performs well on established benchmarks and on previously unsolved problems.

The main contributions of this paper are:
\begin{itemize}
\item We show Delayed Acceptance synthesizes low-level looping programs near the native assembly language level.
\item Using this method, we identify properties of stepping stone programs that enable stochastic search to succeed on challenging program synthesis problems.
\item We show how to solve problems that were previously open challenges, including the ``Collatz Numbers'' benchmark, the problem of synthesizing a fast sorting program without problem-specific background knowledge, and a record-setting
result on a puzzle from the TIS-100 assembly language programming game.
\end{itemize}

\section{Problem Statement}

The program synthesizer receives a set of training examples with input and required output, along with the execution time bound for each.
The synthesizer must return a program in the target language that yields correct output for all the training examples, within the execution time bound.
The returned program is then tested on a set of test examples, and programs that are fully correct on these test examples are reported to {\em generalize perfectly}.
In the problems here, there are several times more test examples than training examples in each problem, and some problems test {\em extrapolation} in which
test examples are much larger than training examples.  All problems here require synthesis of looping programs.

\begin{figure}[t]
\footnotesize{
\begin{Verbatim}[fontfamily=zi4]
delayed_acc_hillclimbing(I):  // I is period length.
  T:=0, B:=0, J:=0            // T is threshold, B is
  Repeat:                     //   best, J is #evals.
    If B==0 Then Z:=rand_pt() // Random init points.
            Else Z:=loc_op(Y) // Local change to Y.
    E:=eval(Z)                // Score (note E>=0).
    If E>=B Then B:=E , N:=Z  // Save N&accept later.
    If E>=T Then Y:=Z         // If E<T undo change.
    J:=J+1 
    If J==I Then:             // End of period:
      If B==T Then Return Y   // No progress; exit.
      Else Y:=N , T:=B , J:=0 // Accept N, update T.
\end{Verbatim}
}
\caption{Delayed Acceptance Hillclimbing.  When applied to program synthesis, N is the period's best program, Y is search's current program, and Z is the new variant to eval.}
\label{boxone}
\end{figure}

\section{Method}

\subsection{Required Format of Target Languages}
\label{formatsection}

Our synthesizer, MAKESPEARE, is intended to target low-level languages, near the native assembly language level.
A program is a sequence of $S$ {\em instructions}.  
Each instruction consists of an {\em opcode} and a single {\em operand}.  
For example, an instruction might have an INC opcode with an operand specifying which variable register is to have its value incremented.
Any of O opcodes can be paired with any of P operands (though some pairings may be nonfunctional).
Specific languages may be instantiated in this format in various ways;
we present two instantiations below.  

Note that native\footnote{We sometimes refer to {\em native} assembly language, to distinguish
from our language which may differ in some respects.} assembly
language instructions often require more than one operand.  The
two instantiations below handle this differently: one
adds a pseudo-opcode to specify the destination operand,
and the other folds one operand into the opcode.  Limiting
our form to one operand may reduce the size of the search space, and increase the probability of search randomly generating a specific required instruction.

In any instantiation, one distinguished opcode $N$ identifies instructions 
that do not execute any function (e.g. NOP no-operation) --
search uses this opcode to simplify solution programs.
In addition, search can generate variant programs by copying operands
from one instruction to another, which benefits from a language in which operands
generally have the same interpretation for different opcodes (although this isn't universally
true for either instantiation below).  Beyond these constraints, search
does not require any further knowledge of the semantics of the language.

\begin{figure}[t]
\footnotesize{
\begin{Verbatim}[fontfamily=zi4]
swapP:=0.1 , doubleP:=0.9 , copyP:=0.5  // parameters
rand_pt(): Return Y with all Y[i] random instructions

loc_op(Y):     // given program Y, make local changes
  If rand(1.0)<swapP Then Z:=swap(Y) // swap 2 instr.
  Else:
    Z:=replacement(Y)
    With probability doubleP, Z:=replacement(Z)
  Return Z

replacement(Y):      // 1 or 2 point replacement in Y
  W := random instruction (random opcode and operand)
  With prob copyP, Set W's opcode:=Y[i]'s for rand i
  With prob copyP, Set W's operand:=Y[j]'s for rand j
  Return Y with Y[k]:=W for random k
\end{Verbatim}
}
\caption{Local Search Operators}
\label{boxtwo}
\end{figure}

\begin{figure}[t]
\footnotesize{
\begin{Verbatim}[fontfamily=zi4,frame=topline]
eval(Y):       // Evaluate program Y on training set
  For each training example X:
    Run Y on X.  Initialize E:=0.
    E:=E+1 if X needs scalar result & Y's is correct
    E:=E+1 per correct output array element
  If fully correct Then: // apply simplicity bonus:
    E:=E+1 per opcode N occurrence in Y 
  Return E
\end{Verbatim}
}
\caption{Eval Scoring Function (see Sec~\protect\ref{tis100section} for TIS-100 detail)}
\label{boxthree}
\end{figure}

\subsection{Search Method}

\begin{table}[t]
\begin{tabular}{|p{3.15in}|}
\hline
{\bf Program:} Sequence of $S$=32 instructions.\\ \hline
{\bf Data:} $R$=6 reg (min $R$ supporting {\bf Input}), memory block \\ \hline
{\bf P Operands:} Operand $x$ is reg $r$, memory $[r]$ if used, immediate constant (0-3),
or jump target $x\lfloor S/P \rfloor$.  Note P=16 if memory is used, P=10 if not (scalar input). \\ \hline
{\bf O=14 Opcodes:} x86-64 opcodes MOV, arithmetic ADD/SUB/IMUL/INC, comparison CMP/TEST, bitwise SHR/SHL, jumps JMP/JZ/JNZ/JG.  
Pseudo-opcode ARG sets current destination; scope continues to next ARG (unaffected by jumps).  See Fig.~\ref{combsort} for example. \\
\hline
{\bf Distinguished Opcode $N$ for Simplification}: ARG. \\
\hline
{\bf Input:} Memory block of $n$ locations with arrays. Initial 
val for $R$ reg: any scalar input, index of last elem of each input array (-1 if none),
first elem of next (if any), and $n$.
With 2d array of row size $m$, 2 reg get values $m$-1 and $m$.\\
\hline
{\bf Output:} Scalar return val is reg R0.
Overwrite input to return  array;
any separate output array must be in input. \\
\hline
{\bf Time and Memory Bounds:} Time bound specified as {\em loopcount}: each backjump taken incurs loopcount 1.
Memory access is limited to memory provided at input.  Time/memory bound violations
terminate execution.  \\ 
\hline
\end{tabular}
\caption{Language Instantiation: Simplified x86-64 Subset}
\label{x86table}
\end{table}

A goal for our work is to use simple stochastic search methods with readily interpreted results, and to explore whether such methods can be effective in searching for looping programs.  To this end, we consider Basic Hillclimbing: local search operators generate small changes
to a current-best candidate solution, and if the changes result in a program at least as good then it is accepted as the current-best candidate.  An immediate problem though is that
Basic Hillclimbing can fairly readily progress by accumulating special-case code along the lines of ``if input=X then output Y''. This can quickly run into a local optimum 
where there's no further room in the $S$ instructions to improve, and in any case such programs will not usually generalize.  In Section~\ref{steppingstones} we observe such behavior for
Basic Hillclimbing.

We therefore turn to {\em Delayed Acceptance hillclimbing} (Figure~\ref{boxone}).  Rather than immediately accepting an update to the current-best candidate solution, we continue to gather additional candidates 
for a {\em period} of $I$ steps, and then accept
the best found during that period.  At that point, the score of the current-best sets a threshold $T$.  During the next period of $I$ steps we take a sequential random walk through candidates having score at least $T$; with large $I$ this random walk can do some global exploration.
We terminate when a period of $I$ steps passes with no improvement.  The resulting search trajectory can be summarized by a small number of milestone programs, one per period.  The method has a single parameter $I$; larger $I$ give longer runs that explore more before accepting improvements.  

We use the name {\em Delayed Acceptance} to place the method in the same family as
previously established {\em step counting hill climbing} \cite{schc} and {\em late acceptance hill climbing} \cite{lahc}.
These are related modifications to hillclimbing, that also use a single parameter like $I$, but differ
in details such as resource management for which our approach is more appropriate to our experiments.  While 
these prior methods were initially developed for domains outside of program synthesis,
late acceptance hillclimbing is competitive with genetic programming on common loop-free benchmarks \cite{gp-lahc}.

\subsubsection{Local Search Operators}

The basic search operation is single-point replacement of one instruction with a random opcode and random operand.
It may be limiting though to restrict to trajectories of single-point changes, each of which must leave a functioning program, so we also
include two-point replacement and swap.  Similar operators have previously been used in stochastic search of programs represented as sequences of instructions \cite{stoke}.

Finally, we use copy operations which modify
replacement by copying operand or opcode from another instruction.  
Operand copying can help when several instructions need the same register operand.
We also note research on 
machine learning to modify the distribution of instructions
selected by an exhaustive search method \cite{deepcoder}; we aren't
exploring such methods here, but copying provides a 
basic way to adapt replacement's instruction distribution.  

The local search operators are detailed in Figure~\ref{boxtwo}.  
Sec.~\ref{parmsection} describes a grid search over parameter settings, and then the
parameters in Fig.~\ref{boxtwo} are fixed for all remaining experiments presented here.

\begin{table}[t]
\begin{tabular}{|p{3.15in}|}
\hline
{\bf Program:} $S$=15 instr.; jump to first
after last finishes. \\ \hline
{\bf Data:} Reg ACC\&BAK hold
integers in $[-999,+999]$. \\ \hline
{\bf P Operands:} Operand $x$ is jump target $\lfloor xS/P \rfloor$ 
or immediate const $\in$ $[-\lfloor P/2 \rfloor,+\lfloor P/2 \rfloor]$. 
Full range: P=1999.  Also tested P=401, P=101, and P=21.
\\ \hline
{\bf O=16 Opcodes:} SAV; SWP; MOV ACC/op,DOWN (sends op to imager); MOV op,ACC; NEG; ADD/SUB op/ACC; NOP; jumps JMP/JEZ/JNZ/JLZ/JGZ. \\ \hline
{\bf Distinguished Opcode $N$ for Simplification}: NOP \\
\hline
{\bf Input:} None needed here.  ACC and BAK are initially 0. \\
\hline
{\bf Output:} 30x18 pixel image.
Imager receives X; Y; seq of colors
at X, X+1, etc.; negative val ends seq.
Program terminates if it exactly gets target image.  Otherwise, score is max (at any point in execution) \# of matching pixels. \\
\hline
{\bf Time Bounds:} 1 cycle/instr., 2 for MOV op,DOWN.\\  
\hline
\end{tabular}
\caption{Language Instantiation: Subset of TIS-100} 
\label{tis100table}
\end{table}

\subsubsection{Evaluating Candidate Programs}

Stochastic search depends on an objective function to be optimized, and the choice of 
objective function affects the existence of partially-correct stepping stones that enable success.
We evaluate candidate programs by running on the training examples and scoring
the output.  Each training example receives 1 point per fully-correct integer in the output (Figure~\ref{boxthree}).  
This is similar to prior approaches \cite{spector} but coarser-grained in that it doesn't
assign partial credit for a partially-correct output integer.

Fully correct programs get a simplicity bonus (\#occurrences of opcode $N$, since it doesn't produce executable code);
this can aid generalization \cite{spectorsimplification}.

Programs run on each training example for its time bound.  
``Time'' is instantiation dependent,
but must be deterministic.  

A run's final simplest program achieving training set success is evaluated on the test set (test set results do not affect scores
used for search).  In experiments here, the primary measure of a run's success is whether
the final program generalizes perfectly, following precedent used for established benchmarks \cite{spector}.

\subsection{Language Instantiations}

We use two instantiations of Sec.~\ref{formatsection}'s language format.
Both include typical assembly language opcodes like MOV
for copying data (e.g. from register to memory), SUB for
subtraction, JMP for jump (goto), and
conditional jumps (e.g. JNZ jumps if previous result was nonzero).

\subsubsection{Simplified x86-64 Subset}

Most of the benchmarks use a language based on a simplified subset of the x86-64 assembly language used in Intel and AMD CPUs (Table ~\ref{x86table}).
This is just a tiny subset of native x86-64, but it does include relevant frequently used native opcodes \cite{freqx86}.

Our focus is limited to programs operating on integers and arrays; extensions such as floating point
and string manipulation would typically be addressed with additional primitives \cite{spectorfull,G3P}.

Time is measured using {\em loopcount}: each backwards jump taken incurs loopcount 1, 
and total loopcount cannot exceed the bound initially specified.  This is a simple deterministic scheme that
usually has an intuitive interpretation; e.g. if a program requires a single pass through an array of $n$ elements
performing constant work on each element, it will use a loopcount of $n$.  Coupled with MAKESPEARE's bonus for simplification,
search favors small loops requiring few iterations. We note though that this is 
a coarse-grained performance measure, and puts no pressure on programs for fine-grained optimizations (e.g. based
on instruction choice and sequencing).  Replacing loopcount with actual execution time measurements
could favor fine-grained optimizations, but would be nondeterministic which we do not support.

\subsubsection{TIS-100}
\label{tis100section}

We explore a small portion of the TIS-100 assembly language programming game \cite{tis100}, that fits easily within our approach (see Table~\ref{tis100table}).
We use just one TIS-100 ``node,'' which leaves aside the game's unusual multi-node parallel programming.
This still allows us to use the game's ``Image Test Pattern'' puzzles, where a
program must generate a 30x18 pixel image to match a target. 

We vary the number of allowed operands P.  The 
full P=1999 constants [-999,+999] have
a low probability
of a randomly-chosen operand hitting a specific needed value (e.g. ``3'' which is a pixel color needed
for benchmarks here).  We therefore also consider restricted ranges with smaller P.

The TIS-100 game scores successful programs according to program size and cycle count.
MAKESPEARE directly optimizes program size, but not cycle count.  We have MAKESPEARE record the lowest
cycle count achieved in each run, at the minimum size correct program that is found.

\subsubsection{Implementation}

Code and data are available for download.\footnote{https://github.com/ChristopherRosin/MAKESPEARE} 
Both languages are compiled to native x86-64 using DynASM~\cite{DynASM}.  For our simplified x86-64 subset,
this gives hardware semantics for native opcodes. Code is instrumented with time/memory bounds checks.  
The full set of Delayed Acceptance experiments below finish in 10 days, using 6 machines, each with a 4-core Intel i7-6700 CPU.

\subsection{Search Parameters} 
\label{parmsection}

With swapP$\in$$\{0,0.1\}$, doubleP$\in$$\{0,0.1,0.5,0.9\}$, copyP $\in$$\{0,0.1,0.5\}$, and I$\in$$\{3k,10k,25k,75k,150k\}$, a grid search was run
on the {\em preliminary benchmarks} in Table~\ref{benchtable}.
For each combination, Delayed Acceptance runs 100 times with max 300k evaluated programs per run.  We find the parameters in Fig.~\ref{boxtwo} with I=75k 
have the most total runs that generalize perfectly, solving all 5 preliminary benchmarks.

Running the grid search with Basic Hillclimbing yields similar results.  In experiments below, we compare Delayed Acceptance and Basic Hillclimbing using the same parameters for both, with Basic Hillclimbing using the same resource bound as Delayed Acceptance and the same stopping criterion (stop after period of length I with no progress).

The choice of I=75k permits only 4 periods with our max 300k evaluated programs per run. Note Delayed Acceptance terminates naturally when progress stops (Fig.~\ref{boxone}), and it usually does so within 10-20 periods.  With unrestricted tests until natural termination at I=75k, on each of the 5 preliminary benchmarks over 98\% of eventual test set successes come within 10 periods -- so we target about 10 periods when using larger numbers of evaluated programs.

Our protocol on a problem starts with 100 runs, each with I=75k and max 4 periods (300k programs), enabling quantitative comparison with other published work on established benchmarks \cite{spector}.  If there's no training set success, the next level is 100 runs with I=2M and max 9 periods (18M programs) each.
If still no training set success, the final level is 30 runs with I=100M and max 10 periods (1 billion programs) each -- this is a reasonable maximum given typical runtimes.  At the first level with training set success, take the simplest training set success program from each run and evaluate on the test set, reporting the total percentage of runs that generalize perfectly.  This protocol works well across a range of difficulty, and is a recommended starting point on new problems.

\begin{table}[t]
\begin{tabular}{| p{2.65in} | p{0.35in} | }
    \hline
{\bf Benchmark} & {\bf Time} \\ \hline\hline
\multicolumn{2}{|l|}{\textbf{\textit{Preliminary Benchmarks}} adapted from {\em \protect\cite{korea}.}}\\ \hline
{\bf Cube Elements}: Given array $a$ of $n$ elements, cube each element (in place) & $2n$ \\ \hline
{\bf 4th Power}: Raise each elem. of $a$ to 4th power & $2n$ \\ \hline
{\bf Sum Sq of Elem}: Given $a$, return $\sum_{i=1}^{n} a[i]^2$ & $2n$ \\ \hline
{\bf Prod Sq of Elem}: Given $a$, return $\prod_{i=1}^{n} a[i]^2$ & $2n$ \\ \hline
{\bf Sum Abs}: Given $a$, return $\sum_{i=1}^{n}$$|a[i]|$ & $2n$ \\ \hline \hline
\multicolumn{2}{|l|}{\textbf{\textit{Established Benchmarks}} {\em \protect\cite{spector}.}}\\ \hline
{\bf Negative To Zero}: Given $a$, $b[i]$$=$$\max(a[i],0)$ & 300 \\ \hline
{\bf Vectors Summed}: Given $a$\&$b$, $c[i]=a[i]+b[i]$ & 300 \\ \hline
{\bf Last Index of Zero}: Return max $i$ with $a[i]=0$ & 300 \\ \hline
{\bf Count Odds}: Given $a$, return count of odd $a[i]$ & 300 \\ \hline
{\bf Mirror Image}: Return 1 iff $a$ is the reverse of $b$ & 300\\ \hline
{\bf Sum of Squares}: Given $x$, return $\sum_{i=1}^{x} i^2$ & 300\\ \hline 
{\bf Collatz Numbers}:  Return \#steps to reach 1 in Collatz sequence starting from input $x$ & 300\\ \hline \hline
\multicolumn{2}{|l|}{\textbf{\textit{Additional/adapted benchmarks.}}} \\ \hline
{\bf Binary Search}: Given $x$ \& sorted $a$, find $a[i]$$=$$x$ & $2\lg n$\\ \hline
{\bf Integer Sqrt}: Given $x$$\geq$$0$, return $\lfloor \sqrt{x} \rfloor$ & $2\lg x$ \\ \hline
{\bf Merge}: Given sorted $a$,$b$, merge into sorted $c$. & $2n$ \\ \hline
{\bf Slow Sort}: Sort $a$ in increasing order (in place) & $2n^{2}$ \\ \hline
{\bf Fast Sort}: Sort $a$ in increasing order (in place) & $2n$$\lg n$$\dagger$ \\ \hline
{\bf Topological Sort}: Given $v$$\times$$v$ edge array $a$, set $b[i]$ to min $L$$\geq$$0$ so that $b[j]$$<$$L$ if $a[j][i]=1$.  & $2n$ \\ \hline
{\bf DAG Sources}: Given $a$ as above, binary $b[i]$=$1$ iff for all $j$, $a[j][i]$=$0$ ($L$=$0$ in Topological Sort). &  $2n$ \\ \hline
\hline
\multicolumn{2}{|l|}{\textbf{\textit{TIS-100 benchmarks}} {\em \protect\cite{tis100}}} \\ \hline
{\bf Image Test Pattern 1}: Set 30x18 img to color 3 & 10000 \\ \hline
{\bf Image Test Pattern 2}: Checkerboard: (X,Y) is color 3 if X+Y is even, else 0 & 10000 \\ \hline
\end{tabular}
\caption{Benchmarks. In Time, $n$ is total input size (may have multiple arrays \& preallocated output space); add 1 to $n$ before taking $\lg$, add 1 to bound before truncating to integer.  $\dagger$ $2n$$\lg n$ Train, and $n^{5/3}$ Test; see Sec.~\protect\ref{combsortsection}. }
\label{benchtable}
\end{table}

\section{Benchmark Descriptions}

Table~\ref{benchtable} describes the benchmarks.  The first 5, used in Sec.~\ref{parmsection},
were the hardest array problems for a constrained exhaustive search \cite{korea} using
input/output examples plus problem-specific program templates (e.g. pre-specifying some loops).
We use only input/output examples, and generate our own larger training and test sets.

The second set are previously established benchmarks \cite{spector},
created partly in response to a community call for stronger benchmarks \cite{betterbenchmarks}.  
These enable comparison with published results \cite{spector,G3P,difficulty,push2018,G3P2018}. 
Appropriate to MAKESPEARE's scope, we use all integer and array benchmarks from this suite which require a loop for MAKESPEARE (excluding string and floating point benchmarks,
which need additional language capabilities \cite{spector,G3P}).
We use the explicit instances made available for download \cite{spectordownload} rather than the protocol for generating new instances \cite{spectorfull}.

The third set includes difficult benchmarks.  Our input/output sets 
require extrapolation (unlike the established benchmarks), with test set array sizes much larger than training.  Time bounds scale
with input size, and extrapolation tests that solution validity and speed scale to large instances.

This third set of benchmarks were adapted from prior work.
Integer Sqrt is a classic benchmark for deductive program synthesis \cite{manna-waldinger-1986}, but hasn't been solved by pure inductive synthesis.
A version of Merge was unsolved in the TerpreT framework for pure inductive synthesis \cite{Terpretarxiv,Terpretworkshop} by several methods including gradient descent, SMT, and {\em SKETCH} \cite{SKETCH}.  A targeted genetic programming effort previously synthesized binary search \cite{sipper2009}.

ADATE uses evolutionary methods and synthesized a slow sort without problem-specific knowledge \cite{inductivesynthesisofrecursive}.
But fast sorts have not been produced by pure inductive synthesis. 
Inductive logic programming \cite{muggleton1994} and genetic
programming \cite{evolvingefficientrecursivesortinglucas} have synthesized
quicksort when given high-level problem-specific primitives like ``partition''.
The neural programming framework 
has learned quicksort using problem-specific program execution
traces which constrain the learned program \cite{dawnsong};
an explicit long-term goal of the work is to remove this constraint.
This same work solved a version of Topological Sort with execution traces.  

DAG Sources is a simpler problem related to Topological Sort.
These graph problems may be hard: our language lacks 2d array indexing;
any needed indexing must be synthesized.

We refer to Merge, Integer Sqrt, Collatz Numbers, Fast Sort, and Topological Sort as {\bf Challenge Problems} with 
no published solution using pure inductive synthesis, despite attention to versions of these in program synthesis literature.

The established benchmarks use up to 200 train and 2000 test examples, and array length up to 50 \cite{spectorfull}.
Other benchmarks use 200 train and 2000 test examples, randomly generated.
These training sets are large compared to programming-by-example work that seeks to minimize numbers of
user-supplied examples in domain-specific applications \cite{gulwanipbe}, but we need more examples for our less-constrained language and 
algorithmic problems.

Most array problems train up to length 21 (length 6 for the first 5 problems) and test extrapolation up to length 2001, with random lengths chosen uniformly. 
Exceptions are the graph problems which train up to $9$$\times$$9$ edge arrays and test up to $201$$\times$$201$, and Binary
Search and Fast Sort which train up to 1001 and 201 (resp.) and test up to 100,001.
Each instance picks random $k$ in [1,63] and then random integers with $k$ bits, except the first 5 benchmarks
pick elements with up to 31 bits and then give them random sign.  Topological Sort instances select random output, then a minimal set of edges that yield that output, and then additional edges with random density.  DAG Sources uses the same graphs as Topological Sort.

The last 2 benchmarks are puzzles from the TIS-100 game.  A program must
output an image matching a target.  This has similarity to other program synthesis
benchmarks requiring output of pictures or patterns 
\cite{patternprograms}.

\addtocounter{footnote}{+1}

\begin{table}[t]
\begin{tabular}{| r | r | r || r | }
    \hline
    \% runs fully & Basic      & {\bf Delayed}    & Smallest \\ 
    successful:   & Hillcl. & {\bf Acceptance} & Program \\ \hline \hline
    \multicolumn{4}{|c|}{First training success in 100 runs of 300k programs} \\ \hline
    {\em Cube Elements} & 44 & {\bf 67} & 5 \\ \hline
    {\em 4th Power} & 99 & {\bf 100} & 4 \\ \hline
    {\em Sum Sq of Elem} & 1 & {\bf 2} & 4 \\ \hline
    {\em Prod Sq of Elem} & {\bf 65} & 36 & 4 \\ \hline
    {\em Sum Abs} & 29 & {\bf 36} & 6 \\ \hline
    Negative To Zero & 42 & {\bf 50} & 6 \\ \hline
    Vectors Summed & 17 & {\bf 26} &  5 \\ \hline
    Last Index of Zero & 88 & {\bf 97} & 4 \\ \hline
    Count Odds & 42 & {\bf 80} & 5 \\ \hline
    Mirror Image & 0 & {\bf 1} & 6 \\ \hline
    Sum of Squares & {\bf 1} & {\bf 1} & 5 \\ \hline\hline
    \multicolumn{4}{|c|}{First training success in 100 runs of 18M programs} \\ \hline
    Slow Sort & 5 & {\bf 60} & 10 \\ \hline 
    Binary Search & 0 & {\bf 17} & 7  \\ \hline
    DAG Sources & 13 & {\bf 59} & 7 \\ \hline 
    {\bf Merge} & 6 & {\bf 13} & 12  \\ \hline
    {\bf Integer Sqrt} & 0 & {\bf 3} &  11  \\ \hline\hline
    \multicolumn{4}{|c|}{First success (or failure) in 30 runs of 1G programs} \\ \hline   
    {\bf Collatz Numbers} & 0 & {\bf 60} & 8 \\ \hline
    {\bf Fast Sort} & 0 & {\bf 3.33} & 14  \\ \hline
    {\bf Topological Sort} & 0 & 0 & - \\ \hline
    \end{tabular}
    \caption[Cap]{MAKESPEARE Delayed Acceptance results, and Basic Hillclimbing comparison: \% of runs with full test set success, at first resource level with any Delayed Acceptance training set success.  {\bf Challenge Problems are shown in bold}; no prior published solution using pure inductive synthesis, despite prior attention in the literature.  {\em Preliminary benchmarks are shown in italics}.  ``Smallest program'' is size (\# non-ARG instructions) of smallest training set success in any run.\addtocounter{footnote}{-1}\protect\footnotemark}
\label{mainresults}
\end{table}

\begin{table}[t]
\begin{tabular}{| r | r | r | r |}
\hline
                    & Push     & G3P      & {\bf MAKESPEARE}  \\ \hline \hline
Negative To Zero    &  82        & {\bf 98} & 50        \\ \hline
Vectors Summed      &  11        & {\bf 85} & 26        \\ \hline
Last Index of Zero  &  72        & 24       & {\bf 97}  \\ \hline
Count Odds          &  12        & 10       & {\bf 80}  \\ \hline
Mirror Image        &  {\bf 100} &  1       & 1         \\ \hline
Sum of Squares      &  {\bf 26}  & 13       & 1         \\ \hline
\end{tabular}
\caption{Comparison: \%runs that generalize perfectly on established benchmarks \cite{spector}, in 100 runs of max 300k programs each.  PushGP \protect\cite{push2018} and G3P \protect\cite{G3P,G3P2018} are genetic programming systems; results shown are best across several published configurations.}
\label{comparisonresults}
\end{table}

\section{Results}

\subsection{Benchmark Results}

Table~\ref{mainresults} shows results (see Sec.~\ref{tis100results} for TIS-100).  All benchmarks including the Challenge Problems are solved by Delayed Acceptance, except Topological Sort.  Inspection of final programs for the easier problems shows they implement an expected algorithm, sometimes in condensed form.  See Sec.~\ref{combsortsection}\&\ref{steppingstones} for example solutions from harder problems.

Table~\ref{mainresults} compares Basic Hillclimbing at the same resource level: it performs worse in nearly all problems, and solves only
one Challenge Problem.

\subsection{Comparison on Established Benchmarks}

The best published results on the established benchmarks are from the well-developed genetic programming systems PushGP \cite{push2018} and G3P \cite{G3P}, which use differing program representations and search operators.  We compare using the train/test protocol and resource bound (100 runs of 300k evaluated programs each) originally established for these benchmarks \cite{spector}.
MAKESPEARE's results are compared in Table~\ref{comparisonresults} to the best results from PushGP and G3P across several published configurations of their methods.
MAKESPEARE has the best result for 2 of the 6 benchmarks, as do each of the other systems.

Beyond genetic programming, other synthesis methods solved less than half of the established benchmarks in a comparative benchmarking effort \cite{difficulty}.

MAKESPEARE has the first reported success on the Collatz Numbers problem \cite{spector,difficulty,GulwaniTestDriveSynthesis}.

\footnotetext{For benchmarks that succeed within 4 periods at I=75k, runs with training set success are extended until natural termination, solely to measure ``smallest program'' (these extensions do not affect other results). 4 periods is rarely enough to simplify programs.}

\subsection{Comparison to Exhaustive Search}

``Smallest program'' in Table~\ref{mainresults} is small enough that one may wonder if exhaustive search could succeed.
But, even with the benefit of constraint to the minimum number of required instructions and registers, we find experimentally that exhaustive search in randomized order
on the smallest example (Last Index of Zero: 4 non-ARG instructions plus one ARG) needs an expected 400M programs until first success: over 1000x as many as MAKESPEARE.  
Other problems need larger programs (non-ARG + ARGs).
Larger exhaustive experiments become prohibitive, and based on search space size the gap (between MAKESPEARE and exhaustive search) could grow rapidly as program sizes increase.

\subsection{Fast Sort Solution}
\label{combsortsection}

\begin{figure}[th]
\footnotesize{
\begin{Verbatim}[fontfamily=zi4]
   ARG R8    
A: MOV 0    // Initialize R8 to 0.
   ARG R2   // R2 is initially n; used as gap size.
   IMUL 3   // Multiply gap size by 3/4.
   SHR 2
   ARG R0   // Initialize R0 to R2; gapped bubble
   MOV R2   //      sort pass compares [R8],[R0].
B: ADD 1    // Ensure final passes' gap is 1 (not 0).
   CMP R6   // R6 is init. n-1 (and doesn't change).
   JG A     // Pass is finished; start another.
   ARG R9    
   MOV [R0] // Compare [R8],[R0]...
   SUB [R8]
   JG C
   ARG [R0] // ...exchange unless [R8]<[R0] already.
   MOV [R8]
   ARG [R8]
   ADD R9
C: INC R8   // Proceed to pass's next pair.
   JNZ B
\end{Verbatim}
}
\caption{MAKESPEARE's Fast Sort (Comb Sort).  Program terminates via time bound.  Note ARG sets destination, with lexical scope down to next ARG (regardless of jumps).}
\label{combsort}
\end{figure}

\begin{figure*}[t]
\centering
\includegraphics[width=0.24\textwidth]{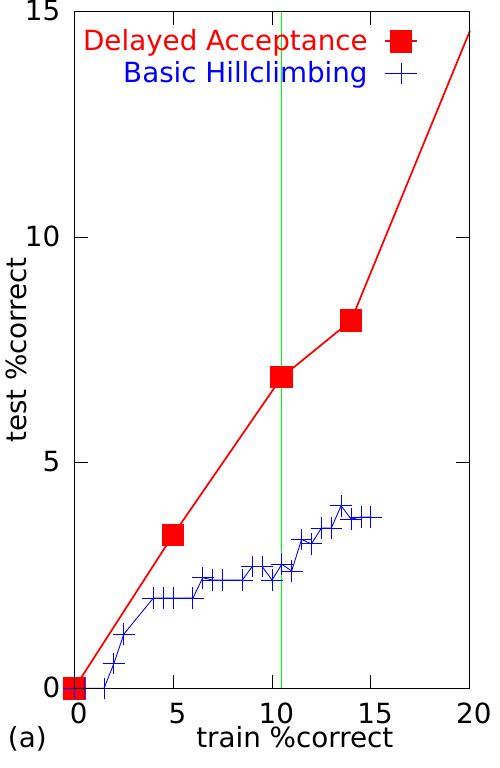} \includegraphics[width=0.37\textwidth]{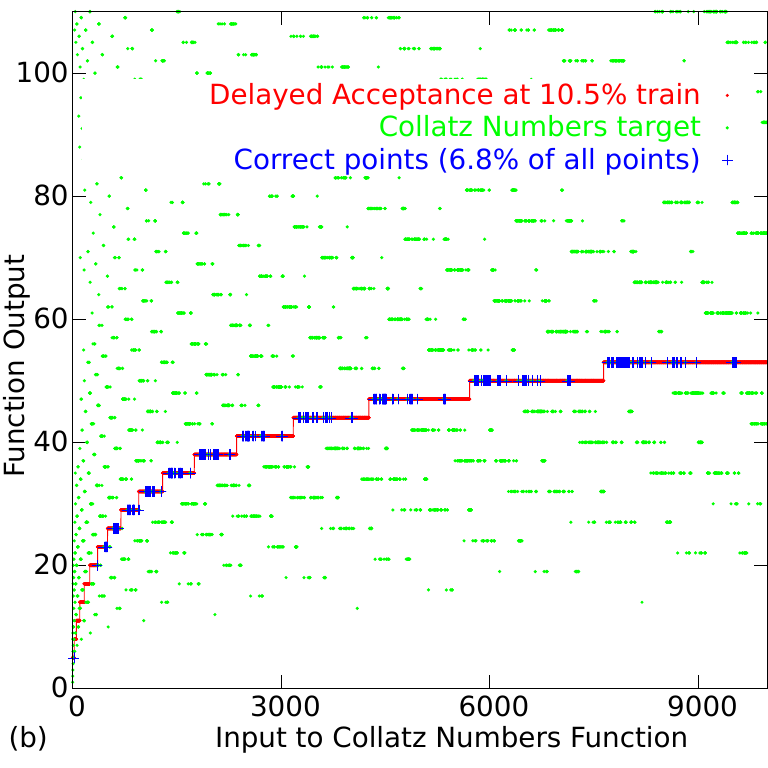} \includegraphics[width=0.37\textwidth]{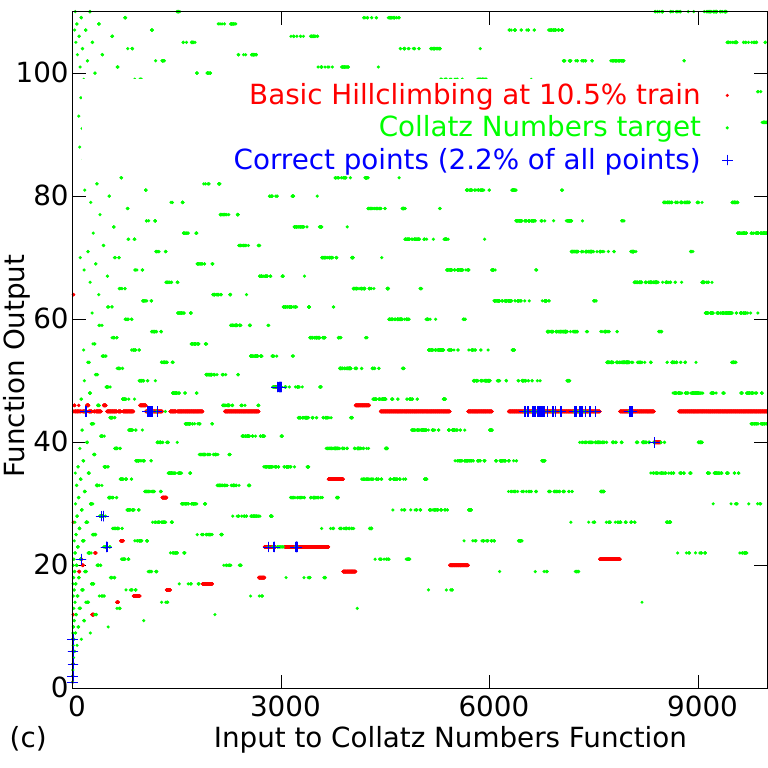} 
\caption{(a) Test \%correct as function of train \%correct, for sample Collatz Numbers trajectories.  Search time goes from left to right too, since train \%correct increases monotonically.  Points show programs accepted by search. Next Delayed Acceptance point, off the graph, is 100\% train\&test. The programs' functions from 10.5\% train (green vertical) are plotted in (b)\&(c).}
\label{traintest}
\end{figure*}

While we obtain multiple Fast Sort training set successes with time bound $2n\lg n$, none of them generalize perfectly to large test examples (up to length 100,001) at the $2n\lg n$ loopcount bound.
We therefore select a relaxed subquadratic bound $n^{5/3}$ that arises in the analysis of particularly simple sorts \cite{sedgewickanalysis}.  
The simplest training set success program (Figure~\ref{combsort}), with 14 non-ARG instructions, generalizes perfectly at this level.

This program turns out to be {\em comb sort}, which was originally suggested by Knuth \cite{knuth}
and later rediscovered independently by others \cite{combsortdobo,combsortbyte}.
Comb sort performs linear-time bubble sort passes on gapped sequences,
multiplying gap size by a constant factor each pass; MAKESPEARE's factor of $3/4$ matches an original choice \cite{combsortdobo}.
When gap size reaches 1, further ungapped passes can continue to complete the sort (MAKESPEARE's version terminates at the time bound).
Comb sort is a variant of Shell sort, which does a complete gapped insertion sort each pass.

While one can craft quadratic-time worst-case inputs for comb sort \cite{quadraticcombsort}, we are unlikely to encounter these in our random inputs.
Comb sort is empirically fast, and even demonstrated faster average times than quicksort on random sequences up to length 1000 (well beyond our training set lengths) \cite{combsortdobo,sedgewickexperiment}.

\subsection{Program Simplification for Generalization}
\label{simplicitysection}

A motivation for program simplification is to improve generalization.  This has been previously observed in genetic programming on benchmarks used here \cite{spectorsimplification}.
We also see a benefit from simplification in our experiments. 

\begin{table}[th]
\begin{tabular}{| r | r |}
\hline
All training success programs within runs & 79.9\% \\ \hline
Final program from each training success run & 84.3\% \\ \hline
Runs yielding Minimal program size & 100.0\% \\ \hline
\end{tabular}
\caption{Percentage of training successes that generalize perfectly.  Averaged over the 18 solved benchmarks in Table~\ref{mainresults}.}
\label{simplificationtable}
\end{table}

Table~\ref{simplificationtable} shows that the ``Final'' simplified program from each training set success run has improved generalization compared to an average over ``All'' accepted training success programs within the run (this trend is only reversed for one of the 18 benchmarks).  The last row shows consistent perfect generalization if we select only the runs achieving training set success with ``Minimal'' program size (achieved within the time bound on max evaluated programs).

``Final program'' is a natural measure of generalization rate, and it varies from 100\% (e.g. Collatz Numbers) down to much lower rates (e.g. 72\% for Merge).
``Minimal program'' shows one approach to encourage reliable generalization is to 
do a large number of runs and take the simplest resulting program.  Another approach (not explicitly tested here) would be to use an independent verification set, to check generalization of programs with training set success.

\subsection{Stepping Stones}
\label{steppingstones}

A {\em stepping stone} here is a partially-correct program that enables search to progress to a complete solution, as opposed to a partially-correct program that leads search to a dead end.

Figure~\ref{traintest}(a) shows typical sample trajectories for Collatz Numbers.
To help illustrate Delayed Acceptance's behavior, we compare it to Basic Hillclimbing.
Delayed Acceptance and Basic Hillclimbing both make initial progress on training and test.  But Basic Hillclimbing then accumulates small special-case training improvements, leading to a complex program representing the piecewise-constant function in Fig.~\ref{traintest}(c).  Such programs do not generalize well, and Basic Hillclimbing's test \%correct stagnates while Delayed Acceptance finds larger training improvements that generalize well and enable progress to a full solution.

The Collatz Numbers target function repeatedly maps $x$ to $\frac{x}{2}$ if even and $3x+1$ if odd, counting steps until $x$ reaches 1.  
This target looks complicated (green in Fig.~\ref{traintest}(b)\&(c)); it isn't obvious that partially-correct stepping stones would exist.

Delayed Acceptance finds a simple partially-correct program that repeatedly multiplies by 3 and divides by 4 while counting steps (Fig.~\ref{traintest}(b)); this generalizes relatively well
and the loop's elements provide a stepping stone towards eventual full solution.  This stepping stone is typical of successful Delayed Acceptance runs.
Final Collatz Numbers solutions resemble the above description of the target function, though the simplest programs don't use the constant $3$ but instead for odd $n$ compute $n + \lceil \frac{n}{2} \rceil$ and count 2 steps.

Other challenging problems, including Integer Sqrt and Binary Search, show typical trajectories similar to the pattern in Fig.~\ref{traintest}(a).  While challenging problems may provide many opportunities for small special-case training improvements, accumulating these can lead to dead ends in terms of generalization and further progress.  Delayed Acceptance bypasses these for more significantly-improved stepping stones that generalize well and enable further progress.

\subsection{Results for TIS-100}
\label{tis100results}

TIS-100 \cite{tis100} is an assembly language programming game that attracted
a dedicated community of players who deeply explored the game and tracked the best programs they could find for each puzzle \cite{tis100leaderboard}.
This enables a concrete comparison of synthesis with strong human efforts, but
we are not aware of previous strong/published program synthesis results on TIS-100.

The TIS-100 game evaluates solution programs by three measures: number of instructions, number of cycles, and number of ``nodes'' (which is always 1 in our implementation).  
For each benchmark, the community leaderboard tracks the best known programs by these measures \cite{tis100leaderboard}. 
MAKESPEARE directly minimizes the number of instructions, so we focus on that category.

For Image Test Pattern 1 with 100 runs at I=2M, MAKESPEARE matches the best reported program at 7 instructions and 2282 cycles, with the same approach of a nested loop using just a single index \cite{imagetestpattern1reddit}.  
Synthesis is successful despite a wide range of P=1999 operands, with the final
program utilizing a large constant.

For Image Test Pattern 2, the community had no prior report of a single-node solution with under 11 instructions.
To explore a variety of solutions, and to see whether restricted ranges of operands could help, we performed extended sets of 5000 runs at I=2M 
with each of the 4 operand ranges in Table~\ref{tis100table}.  The best program was found with P=401 and has 9 instructions and 3596 cycles, which sufficed to set a record and claim a spot on the leaderboard \cite{tis100leaderboard}.

\section{Conclusion}

We have shown that a simple Delayed Acceptance hillclimbing method successfully synthesizes low-level looping programs, near the assembly language level.  
Delayed Acceptance bypasses small gains to identify significantly-improved stepping stone programs that tend to generalize and enable further progress. 
Novel results include (a) the first reported solution of the ``Collatz Numbers'' benchmark, (b) in the problem of fast sorting, the emergence of comb sort, an empirically fast sort; this is
a significant milestone in the long-term goal of synthesizing efficient sorting algorithms from low-level primitives, and (c) algorithmic novelty in the form of
a record-setting program in the TIS-100 assembly language programming game.  

\section{Acknowledgments}

Thanks to Mark Land and anonymous reviewers for helpful comments.

\bibliographystyle{aaai}
\bibliography{pr}

\begin{thebibliography}{}

\bibitem[\protect\citeauthoryear{Agapitos and
  Lucas}{2006}]{evolvingefficientrecursivesortinglucas}
Agapitos, A., and Lucas, S.
\newblock 2006.
\newblock Evolving efficient recursive sorting algorithms.
\newblock In {\em IEEE CEC},  2677--2684.

\bibitem[\protect\citeauthoryear{Balog \bgroup et al\mbox.\egroup
  }{2017}]{deepcoder}
Balog, M.; Gaunt, A.; Brockschmidt, M.; Nowozin, S.; and Tarlow, D.
\newblock 2017.
\newblock Deep{C}oder.
\newblock In {\em ICLR}.

\bibitem[\protect\citeauthoryear{Burke and Bykov}{2008}]{lahc}
Burke, E., and Bykov, Y.
\newblock 2008.
\newblock A late acceptance strategy in hill-climbing for exam timetabling
  problems.
\newblock In {\em PATAT}.

\bibitem[\protect\citeauthoryear{Bykov and Petrovic}{2016}]{schc}
Bykov, Y., and Petrovic, S.
\newblock 2016.
\newblock A step counting hill climbing algorithm applied to university
  examination timetabling.
\newblock {\em J. Sched.} 19:479--492.

\bibitem[\protect\citeauthoryear{Cai, Shin, and Song}{2017}]{dawnsong}
Cai, J.; Shin, R.; and Song, D.
\newblock 2017.
\newblock Making neural programming architectures generalize via recursion.
\newblock In {\em ICLR}.

\bibitem[\protect\citeauthoryear{Devlin \bgroup et al\mbox.\egroup
  }{2017}]{robustfill}
Devlin, J.; Uesato, J.; Bhupatiraju, S.; Singh, R.; Mohamed, A.; and Kohli, P.
\newblock 2017.
\newblock Robust{F}ill: neural program learning under noise {I}/{O}.
\newblock In {\em ICML}.

\bibitem[\protect\citeauthoryear{Dobosiewicz}{1980}]{combsortdobo}
Dobosiewicz, W.
\newblock 1980.
\newblock An efficient variation of bubble sort.
\newblock {\em Inf Proc Lett} 11:5--6.

\bibitem[\protect\citeauthoryear{Drozdek}{2005}]{quadraticcombsort}
Drozdek, A.
\newblock 2005.
\newblock Worst case for comb sort.
\newblock {\em Informatyka Teoretyczna i Stosowana} 5:23--27.

\bibitem[\protect\citeauthoryear{Feser, Chaudhuri, and Dillig}{2015}]{pldi2015}
Feser, J.; Chaudhuri, S.; and Dillig, I.
\newblock 2015.
\newblock Synthesizing data structure transformations from input-output
  examples.
\newblock In {\em PLDI}.

\bibitem[\protect\citeauthoryear{Forstenlechner \bgroup et al\mbox.\egroup
  }{2017}]{G3P}
Forstenlechner, S.; Fagan, D.; Nicolau, M.; and O'Neill, M.
\newblock 2017.
\newblock A grammar design pattern for arbitrary program synthesis problems in
  {G}{P}.
\newblock In {\em Evostar}.

\bibitem[\protect\citeauthoryear{Forstenlechner \bgroup et al\mbox.\egroup
  }{2018}]{G3P2018}
Forstenlechner, S.; Fagan, D.; Nicolau, M.; and O'Neill, M.
\newblock 2018.
\newblock Towards effective semantic operators for program synthesis in genetic
  programming.
\newblock In {\em GECCO}.

\bibitem[\protect\citeauthoryear{Gaunt \bgroup et al\mbox.\egroup
  }{2016a}]{Terpretworkshop}
Gaunt, A.; Brockschmidt, M.; Singh, R.; Kushman, N.; Kohli, P.; Taylor, J.; and
  Tarlow, D.
\newblock 2016a.
\newblock Summary -- {T}erpret.
\newblock {\em NIPS NAMPI Workshop, arXiv:1612.00817}.

\bibitem[\protect\citeauthoryear{Gaunt \bgroup et al\mbox.\egroup
  }{2016b}]{Terpretarxiv}
Gaunt, A.; Brockschmidt, M.; Singh, R.; Kushman, N.; Kohli, P.; Taylor, J.; and
  Tarlow, D.
\newblock 2016b.
\newblock {T}erpret.
\newblock {\em arXiv:1608.04428}.

\bibitem[\protect\citeauthoryear{GltyBystndr}{2016}]{imagetestpattern1reddit}
GltyBystndr.
\newblock 2016.
\newblock Image test pattern 1:2282/1/7.
\newblock www.reddit.com/r/tis100/comments/3ab1t7/.

\bibitem[\protect\citeauthoryear{Gulwani \bgroup et al\mbox.\egroup
  }{2011}]{gulwaniloopfree}
Gulwani, S.; Jha, S.; Tiwari, A.; and Venkatesan, R.
\newblock 2011.
\newblock Synthesis of loop-free programs.
\newblock In {\em PLDI}.

\bibitem[\protect\citeauthoryear{Gulwani, Polozov, and
  Singh}{2017}]{gulwanibook}
Gulwani, S.; Polozov, O.; and Singh, R.
\newblock 2017.
\newblock Program synthesis.
\newblock {\em Foundations \& Trends in Programming Languages} 4:1--119.

\bibitem[\protect\citeauthoryear{Gulwani}{2011}]{flashfill}
Gulwani, S.
\newblock 2011.
\newblock Automating string processing in spreadsheets using {I}/{O} examples.
\newblock In {\em PoPL}.

\bibitem[\protect\citeauthoryear{Helmuth and Spector}{2015a}]{spectordownload}
Helmuth, T., and Spector, L.
\newblock 2015a.
\newblock General program synthesis training and test examples {C}{S}{V}s.
\newblock github.com/thelmuth/Program-Synthesis-Benchmark-Data.

\bibitem[\protect\citeauthoryear{Helmuth and Spector}{2015b}]{spector}
Helmuth, T., and Spector, L.
\newblock 2015b.
\newblock General program synthesis benchmark suite.
\newblock In {\em GECCO},  1039--1046.

\bibitem[\protect\citeauthoryear{Helmuth and Spector}{2015c}]{spectorfull}
Helmuth, T., and Spector, L.
\newblock 2015c.
\newblock Detailed problem descriptions for general program synthesis benchmark
  suite.
\newblock web.cs.umass.edu/publication/details.php?id=2387.

\bibitem[\protect\citeauthoryear{Helmuth \bgroup et al\mbox.\egroup
  }{2017}]{spectorsimplification}
Helmuth, T.; McPhee, N.; Pantridge, E.; and Spector, L.
\newblock 2017.
\newblock Improving generalization of evolved programs through automatic
  simplification.
\newblock In {\em GECCO}.

\bibitem[\protect\citeauthoryear{Helmuth, McPhee, and Spector}{2018}]{push2018}
Helmuth, T.; McPhee, N.; and Spector, L.
\newblock 2018.
\newblock Program synthesis using uniform mutation by addition and deletion.
\newblock In {\em GECCO}.

\bibitem[\protect\citeauthoryear{Hofmann \bgroup et al\mbox.\egroup
  }{2007}]{inductivesynthesisofrecursive}
Hofmann, M.; Hirschberger, A.; Kitzelmann, E.; and Schmid, U.
\newblock 2007.
\newblock Inductive synthesis of recursive functional programs.
\newblock In {\em KI},  468--472.

\bibitem[\protect\citeauthoryear{Incerpi and
  Sedgewick}{1987}]{sedgewickexperiment}
Incerpi, J., and Sedgewick, R.
\newblock 1987.
\newblock Practical variations of shellsort.
\newblock {\em Inf Proc Lett} 26:37--43.

\bibitem[\protect\citeauthoryear{Knuth}{1973}]{knuth}
Knuth, D.
\newblock 1973.
\newblock {\em The {A}rt of {C}omputer {P}rogramming, vol. 3}.
\newblock Reading, MA, Addison-Wesley.

\bibitem[\protect\citeauthoryear{Lacey and Box}{1991}]{combsortbyte}
Lacey, S., and Box, R.
\newblock 1991.
\newblock A fast, easy sort.
\newblock {\em B{Y}{T}{E}} 4:315.

\bibitem[\protect\citeauthoryear{Lawlor}{2012}]{freqx86}
Lawlor, O.
\newblock 2012.
\newblock Instruction encoding \& frequency.
\newblock https://bit.ly/2PPpLFq.

\bibitem[\protect\citeauthoryear{Leaderboard}{2018}]{tis100leaderboard}
Leaderboard.
\newblock 2018.
\newblock www.reddit.com/r/tis100/wiki/index.

\bibitem[\protect\citeauthoryear{Manna and Waldinger}{1980}]{deductive}
Manna, Z., and Waldinger, R.
\newblock 1980.
\newblock A deductive approach to program synthesis.
\newblock In {\em ACM TOPLAS}.

\bibitem[\protect\citeauthoryear{Manna and
  Waldinger}{1986}]{manna-waldinger-1986}
Manna, Z., and Waldinger, R.
\newblock 1986.
\newblock The origin of a binary-search paradigm.
\newblock SRI Technical Note 351R.

\bibitem[\protect\citeauthoryear{McDermott and Nicolau}{2017}]{gp-lahc}
McDermott, J., and Nicolau, M.
\newblock 2017.
\newblock Late-acceptance hill-climbing with a grammatical program
  representation.
\newblock In {\em GECCO}.

\bibitem[\protect\citeauthoryear{McDermott \bgroup et al\mbox.\egroup
  }{2012}]{betterbenchmarks}
McDermott, J.; White, D.; Luke, S.; Manzoni, L.; Castelli, M.; Vanneschi, L.;
  Jaskowski, W.; Krawiec, K.; Harper, R.; De~Jong, K.; and O'Reilly, U.-M.
\newblock 2012.
\newblock Genetic programming needs better benchmarks.
\newblock In {\em GECCO},  791--798.

\bibitem[\protect\citeauthoryear{Muggleton and De~Raedt}{1994}]{muggleton1994}
Muggleton, S., and De~Raedt, L.
\newblock 1994.
\newblock Inductive logic programming.
\newblock {\em J Logic Prog} 19:629--679.

\bibitem[\protect\citeauthoryear{Pall}{2017}]{DynASM}
Pall, M.
\newblock 2017.
\newblock Dynasm.
\newblock luajit.org/dynasm.html.

\bibitem[\protect\citeauthoryear{Pantridge \bgroup et al\mbox.\egroup
  }{2017}]{difficulty}
Pantridge, E.; Helmuth, T.; McPhee, N.; and Spector, L.
\newblock 2017.
\newblock On the difficulty of benchmarking inductive program synthesis
  methods.
\newblock In {\em GECCO}.

\bibitem[\protect\citeauthoryear{Perelman \bgroup et al\mbox.\egroup
  }{2014}]{GulwaniTestDriveSynthesis}
Perelman, D.; Gulwani, S.; Grossman, D.; and Provost, P.
\newblock 2014.
\newblock Test-driven synthesis.
\newblock In {\em PLDI}.

\bibitem[\protect\citeauthoryear{Raza, Gulwani, and
  Milic-Frayling}{2014}]{gulwanipbe}
Raza, M.; Gulwani, S.; and Milic-Frayling, N.
\newblock 2014.
\newblock Programming by example using least general generalizations.
\newblock In {\em AAAI}.

\bibitem[\protect\citeauthoryear{Schkufza, Sharma, and Aiken}{2016}]{stoke}
Schkufza, E.; Sharma, R.; and Aiken, A.
\newblock 2016.
\newblock Stochastic program optimization.
\newblock {\em Comm. ACM} 59:114--122.

\bibitem[\protect\citeauthoryear{Sedgewick}{1996}]{sedgewickanalysis}
Sedgewick, R.
\newblock 1996.
\newblock Analysis of {S}hellsort and related algorithms.
\newblock In {\em ESA}.

\bibitem[\protect\citeauthoryear{So and Oh}{2017}]{korea}
So, S., and Oh, H.
\newblock 2017.
\newblock Synthesizing imperative programs for introductory programming
  assignments.
\newblock In {\em Static Analysis Symposium}.

\bibitem[\protect\citeauthoryear{So and Oh}{2018}]{patternprograms}
So, S., and Oh, H.
\newblock 2018.
\newblock Synthesizing pattern programs from examples.
\newblock In {\em IJCAI}.

\bibitem[\protect\citeauthoryear{Solar-Lezama}{2008}]{SKETCH}
Solar-Lezama, A.
\newblock 2008.
\newblock {\em Program synthesis by sketching}.
\newblock Ph.D. Dissertation, UC Berkeley.

\bibitem[\protect\citeauthoryear{Waldinger and Lee}{1969}]{prowijcai69}
Waldinger, R., and Lee, R.
\newblock 1969.
\newblock {P}{R}{O}{W}: A step toward automatic program writing.
\newblock In {\em IJCAI 1}.

\bibitem[\protect\citeauthoryear{Wolfson and Sipper}{2009}]{sipper2009}
Wolfson, K., and Sipper, M.
\newblock 2009.
\newblock Evolving efficient list search algorithms.
\newblock In {\em Evolution Artificielle (EA)}.

\bibitem[\protect\citeauthoryear{Yin and Neubig}{2017}]{cardgamebenchmark}
Yin, P., and Neubig, G.
\newblock 2017.
\newblock A syntactic neural model for general-purpose code generation.
\newblock In {\em ACL}.

\bibitem[\protect\citeauthoryear{Zachtronics}{2015}]{tis100}
Zachtronics.
\newblock 2015.
\newblock {T}{I}{S}-100.

\end{thebibliography}

\end{document}